# Real World Robustness from Systematic Noise

Yan Wang
SenseTime Research
Beijing, Beijing, China
wangyan3@sensetime.com

Yuhang Li
Yale University
New Haven, CT, USA
yuhang.li@yale.edu

Ruihao Gong*
SenseTime Research & Beihang
University
Beijing, Beijing, China
gongruihao@sensetime.com

Tianzi Xiao
SenseTime Research
Beijing, Beijing, China
danache0405@gmail.com

Fengwei Yu
SenseTime Research
Beijing, Beijing, China
yufengwei@sensetime.com

## ABSTRACT

Systematic error, which is not determined by chance, often refers to the inaccuracy (involving either the observation or measurement process) inherent to a system. In this paper, we exhibit some long-neglected but frequent-happening adversarial examples caused by systematic error. More specifically, we find the trained neural network classifier can be fooled by inconsistent implementations of image decoding and resize. This tiny difference between these implementations often causes an accuracy drop from training to deployment. To benchmark these real-world adversarial examples, we propose ImageNet-S dataset, which enables researchers to measure a classifier's robustness to systematic error. For example, we find a normal ResNet-50 trained on ImageNet can have 1%~5% accuracy difference due to the systematic error. Together our evaluation and dataset may aid future work toward real-world robustness and practical generalization.

## 1 INTRODUCTION

Recently deep learning has shown remarkable success in image recognition [14, 17, 25], speech recognition [13, 19] and natural language processing [4, 8]. However, their security and robustness are greatly challenged by the delicately designed adversarial attacks [6, 12, 20]. The adversarial examples usually fool the neural network without significant visual differences, demonstrating the vulnerability of deep neural networks. This phenomenon attracts broad research interests for different kinds of attacks and defenses.

Currently, the mainstream attack noises can be categorized into adversarial noises and natural noises. The adversarial noises need to be generated according to the model's specific information such as gradients calculated in backpropagation. Thus, they often achieve satisfactory attack success rates but are rare in real-world scenarios due to the model dependency. In the contrast, there exist natural noises such as blur or corruption [15, 16] that are agnostic to model and frequently encountered in the natural scenario. Although they have less damage on model robustness compared with adversarial noises, it is impossible to overlook them because of the border range of influence in nature. Therefore, evaluating the robustness of both types of noises is necessary. Some existing defense work [18] surprisingly finds that deep neural networks optimized by adversarial training might be sensitive to natural noises. This indicates that we need diverse noises for a comprehensive understanding of the model's robustness.

However, the commonly adopted noises are not diverse enough. Besides the man-made noise and natural noise, we point out a new kind of noise-induced by the inconsistent resize or decoder implementation widely existing in the inference system of deep learning model. To be specific, we call it *systematic noise* since they are not corresponded to the model or the input image, but is only related to the inference software and hardware system. Once the software or hardware-software changes, the noise will occur. For example, we always utilize the Pillow package for image decoding and resizing in the PyTorch [21] training system. But when it is deployed on the edge device, it may not support Pillow and we have to seek an alternative for the practical deployment. Sometimes due to the hardware and software's restriction, the alternative has no way to keep alignment with the origin Pillow implementation, leading to an inevitable systematic noise. The maximum perturbation caused by this systematic noise is only a 1-pixel value but may introduce an obvious accuracy drop (about 1% ~ 5%) in our experiment. Different from the extensively studied adversarial noises and natural noises, this kind of systematic noise is long-term neglected and needs to be included in the robustness evaluation.

In order to attract attention to systematic noise, we summarize the major systematic error sources in the deep learning system and propose a corresponding ImageNet-S dataset, providing provides 3 kinds of decoder noises and 6 kinds of resize noises. It also supports the flexible custom extension by users. We conduct a pioneering benchmark for some prevalent neural architectures on the systematic robustness and find some useful insights: (1) The effect of using different decoders is small while it may have a big impact when using different resize methods; (2) Usually, models in the same architecture will have better robustness of resize with FLOPs (Floating Point Operations) of it get larger, but it doesn't work for the robustness of decoder; (3) Adversarial methods can increase the robustness of resize method, but we need to find a balance between robustness and clean accuracy. Along with the ImageNet-S benchmark dataset, we also propose a simple yet effective mixed training strategy to improve systematic robustness. Our contributions can be summarized as below:

- We firstly indicate a kind of long-neglected but frequent-happening systematic noise caused by the inconsistent deploy software and hardware environment. It may harm the model's performance in practical application.
- To comprehensively evaluate the effect of systematic noise, we construct an ImageNet-S dataset, which consists of 6 different kinds of resize functions and 3 different kinds of decoder functions. We benchmark the robustness of the prevalent ResNet, RegNetX, MobileNet-V2 networks, and "robust" networks with adversarial training. Even though the systematic noises are slight, they may introduce an up to 2% accuracy jitter, especially for the resize noise.
- To improve the stability when facing systematic noise, we propose a simple yet effective mix training strategy. It can be easily applied in the training routine without much effort but enable the network to enjoy a consistent performance under different deployment systems.

---

*Corresponding Author.

## 2 RELATED WORK

**Adversarial Examples.** The adversarial examples were first introduced by [26]. From then on, various adversarial attacks were proposed, such as FGSM [12], PGD [20], C&W [5] and some black-box attack methods [2]. Driven by the emergence of adversarial noises, corresponding defense techniques also arose, including adversarial training [10, 24, 28], data augmentation [11] and regularization [9]. The adversarial examples are always dependent on the model to attack, especially for the block-box attack. Thus they suffer a low transferability and rarely occur in the practical scenario.

**Natural Noises.** Besides the adversarial examples, the community also realizes the importance of natural noises that are widely existing in the real world. Some representative datasets are constructed to simulate the natural noise, such as ImageNet-P, ImageNet-C [15], and ImageNet-A, ImageNet-O [16]. These noises are model-agnostic and may cause perceptible perturbation. Natural noises such as Snow noise and Frost noise can measure the robustness of a model in the wild.

## 3 ADVERSARIAL CORRUPTIONS FROM SYSTEMATIC ERROR

In this section, we outline the different system implementations for image decoding and resize. Then, we introduce our ImageNet-S dataset to benchmark the robustness from the different sources of decoding and resizing.

### 3.1 Notations

Let us denote a RGB image tensor as $X \in [0,1]^{w \times h \times 3}$ where $w$ and $h$ are the image width and height, respectively. RGB images have 3 channels standing for red, green, and blue colors. However, in computer systems, the images are not stored in this format. Typically there is a raw file (e.g. JPEG) which encodes the tensor $X$ to $V$. In the case of JEPG encoding, $X$ will be processed by color space conversion, discrete cosine transform (DCT), quantization, and Huffman encoding to $V$. For image decoding, we should reconstruct the image tensor $X$ from $V$. As for image resize, the operation can be represented by $\text{Resize}(X) : [0,1]^{w \times h \times 3} \to [0,1]^{w' \times h' \times 3}$, where $w'$ and $h'$ are the resized width and height, respectively.

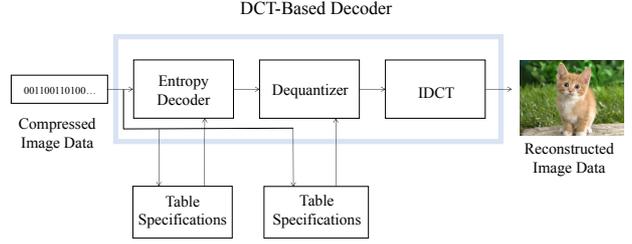

Figure 1: DCT-Based Decoder Processing Steps.

Generally, in neural network inference, the image is processed by decoding and resize ($V \to X \to X'$)[1]. With the different implementations of these two operations, output $X'$ can have different values, which will be discussed in the next two sections.

### 3.2 Image Decoding

Image decoding refers to the process of translating the raw file (e.g. JPEG) back to an RGB or YUV 3 channel image tensor ($V \to X$). Image decoding can be divided into several steps: (1) read the raw file from disk as bytes, (2) variable-length decoding (3) zigzag scan, (3) dequantization, (4) inverse discrete cosine transform (iDCT), (5) color conversion and reorder, shown as Figure 1. The fourth step iDCT occupies the majority of the computational cost in the decoding process, which is given by:

$$f[m,n] = \sum_{k=0}^{N-1} \sum_{l=0}^{N-1} \alpha(k)\alpha(l)F(k,l) \\ cos[\frac{(2m+1)\pi k}{2N}]cos[\frac{(2n+1)\pi l}{2N}] \quad (1)$$

where,

$$\alpha(k) \, and \, \alpha(l) = \begin{cases} \sqrt{\frac{1}{N}} \, if \, k = 0 \\ \sqrt{\frac{2}{N}} \, if \, k \neq 0 \end{cases}$$

In theory, the principle of this process is fixed, but we find decoding one image file in different third-party libraries (e.g. OpenCV [3], Pillow [29], FFmpeg [27]) will output different RGB tensors. This is because they have a unique self-implemented tool to decode the images, especially the iDCT step. We find some libraries prefer to use fast inverse discrete cosine transform (Fast iDCT) instead of the vanilla one, which sacrifices the image quality for the decoding speed. Furthermore, there would be some minor errors in the implementation, such as (ADD EXAMPLES.) These minor errors can cause a shift in the pixel values of the final RGB tensor. Eventually, a single difference in pixel value could fool the neural network to change its prediction. Most of the time, when changing the decoding tools used in training to another one in inference, we observe a drop in accuracy.

---

[1]Sometimes the center crop operation is also utilized. It has a similar impact with the resize operation, therefore we omit its discussion in this paper.

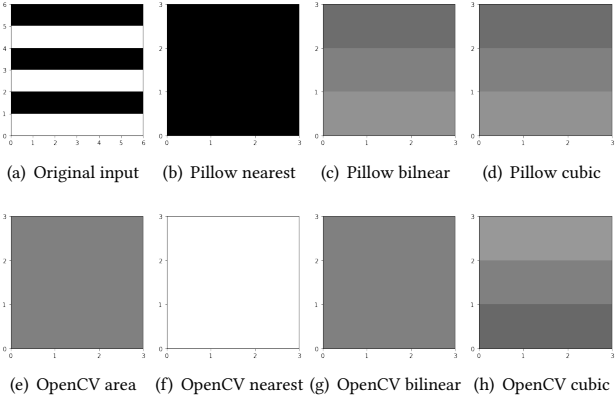

(a) Original input  (b) Pillow nearest  (c) Pillow bilnear  (d) Pillow cubic

(e) OpenCV area  (f) OpenCV nearest  (g) OpenCV bilinear  (h) OpenCV cubic

Figure 2: Difference between multi resize methods and tools. Here we visualize the resize output from a synthetic $6 \times 6$ image. The first row and the second row contain the Pillow and OpenCV implementation, respectively. We observe the difference not only at the algorithm level but also at the library level.

To evaluate the influence of different image decoders, in Section 4, we will compare with three different decoder tools including OpenCV, Pillow and FFmpeg which are widely used in the field of computer vision.

## 3.3 Image Resize Method

Image resize is adopted to scale the image resolution to a different size, either up (increase resolution) or down (decrease resolution). In resize operation, one needs to predict the pixel value at an unseen position. This is often performed by different *interpolation* algorithms. We summarize several common algorithms, like nearest neighbor, bilinear, bicubic, etc.

*3.3.1 Nearest-neighbor Interpolation.* This algorithm, also known as *Proximal Interpolation*, directly selects the value of its nearest known pixel. This mode does not consider the relative magnitude of its neighbor pixels. Therefore, it can preserve sharp details in pixel art, but also introduce jaggedness in previously smooth images [1]. Mathematically, for 2D images where spatial coordinate is represented by $(x, y)$, the nearest neighbor interpolation will find the existing pixel with lowest Euclidean distance i.e., $X[\arg\min_{x,y}((x - x')^2 + (y - y')^2)]$.

*3.3.2 Bilinear Interpolation.* Bilinear interpolation is a simple extension from linear interpolation by operating on the 2D variables. Unlike Nearest-neighbor algorithm, this mode consider the value of its neighbor and has a rather continuous interpolation effect. Mathematically, our objective is to predict the value of the function $f(\cdot)$ at the unknown point $(x, y)$. Assume we have four near coordinates: $Q_{11} = (x1, y1)$, $Q_{12} = (x1, y2)$, $Q_{21} = (x2, y1)$, and $Q_{22} = (x2, y2)$. Their values are already know, for example $f(Q_{11})$. The formulation of bilinear interpolation is given by:

$$f(x,y) = \frac{y_2 - y}{y_2 - y_1} f(x, y_1) + \frac{y - y_1}{y_2 - y_1} f(x, y_2), \quad (2)$$

Table 1: Image Resize Tools and Their Method.

|  | Nearest | Bilinear | Cubic | Lanczos | Area | Box | Hamming |
|---|---|---|---|---|---|---|---|
| OpenCV | ✓ | ✓ | ✓ | ✓ | ✓ | ✗ | ✗ |
| Pillow | ✓ | ✓ | ✓ | ✓ | ✗ | ✓ | ✓ |

where,

$$f(x, y_1) = \frac{x_2 - x}{x_2 - x_1} f(Q_{11}) + \frac{x - x_1}{x_2 - x_1} f(Q_{21}),$$
$$f(x, y_2) = \frac{x_2 - x}{x_2 - x_1} f(Q_{12}) + \frac{x - x_1}{x_2 - x_1} f(Q_{22}).$$

*3.3.3 Bicubic Interpolation.* In contrast to the bilinear interpolation which only takes 4 pixels ($2 \times 2$), the bicubic interpolation takes 16 pixels ($4\times4$). This method is even more smoothed than bilinear since it considers more neighbors. The algorithm tries to use existing known pixel values to fit a binary cubic function

$$f(x, y) = \sum_{i=0}^{3}\sum_{j=0}^{3} a_{ij} x^i y^j \quad (3)$$

To find the total 16 coefficients $a_{ij}, ij \in \{0, 1, 2, 3\}$, we need to solve a system of linear equations $A\alpha = x$. Due to the complexity of this algorithm, we refer the readers to this link[2] for more details. Bicubic interpolation yields better performance than the previous two algorithms, however, it also needs huge time to solve the linear equations to find optimal interpolated values.

## 3.4 Image Resize Library

We summarize two most commonly used image resize tools in the field of computer vision. Table 1 shows the image resize tools and their corresponding resize method.

Similar to the image decoder, the principal of one kind of image resize method is basically the same, the result will be different due to the difference between implementation detail. Figure 2 shows the difference between resize tools using the same resize method. From this figure, we can see that the gap between different tools using the same method is not small. In the real-world images, different resize tools applying on the same image may not lead to an obvious distance between results, but the gap between them can result in a destabilization of model robustness.

## 3.5 Visualization of Resize Algorithms

Apart from our mentioned algorithms, many more exists. Here we do not introduce them one by one. To provide an intuitive understanding of the difference, we visualize the results of various resize methods in Figure 2. We use a special $6 \times 6$ synthetic image as the original image to magnify the effect of resize. All images are resized to $3\times 3$ resolution. Other interpolation methods like *lanczos, area* are also visualized.

Due to the specific characteristics of this synthetic image, we observe a huge disparity between these resize methods. Take the nearest neighbor as an example, this method results in pure color resized images, either whole black (Pillow) or whole white (OpenCV),

---
[2] https://www.ece.mcmaster.ca/~xwu/interp_1.pdf

Table 2: Top-1 accuracy of various pre-trained network architectures on ImageNet-S. "N, L, C" refer to nearest neighbor, bilinear, and bicubic resize method. DALI is excluded when calculating the mean and standard deviation.

| Architecture | Decode | | | | | Resize | | | | | | |
|---|---|---|---|---|---|---|---|---|---|---|---|---|
| | DALI | OpenCV | Pillow | FFmpeg | Mean ± Std. | Pillow-N | Pillow-L | Pillow-C | OpenCV-N | OpenCV-L | OpenCV-C | Mean ± Std. |
| ResNet-18 | 70.860 | 69.720 | 69.718 | 69.716 | 69.718 ± 2.00E-03 | 68.906 | 69.720 | 70.132 | 68.842 | 70.396 | 70.094 | 59.728 ± 6.62E-01 |
| ResNet-34 | 75.010 | 74.182 | 74.176 | 74.176 | 74.178 ± 3.46E-03 | 73.488 | 74.182 | 74.490 | 73.428 | 74.834 | 74.538 | 63.567 ± 5.82E-01 |
| ResNet-50 | 78.140 | 77.398 | 77.380 | 77.408 | 77.395 ± 1.42E-02 | 76.764 | 77.398 | 77.736 | 76.728 | 77.996 | 77.858 | 66.355 ± 5.54E-01 |
| ResNet-101 | 79.770 | 79.110 | 79.094 | 79.102 | 79.102 ± 8.00E-03 | 78.386 | 79.112 | 79.470 | 78.660 | 79.544 | 79.314 | 67.785 ± 4.65E-01 |
| MobileNetV2-0.5 | 64.938 | 62.942 | 62.962 | 62.976 | 62.960 ± 1.71E-02 | 61.800 | 62.942 | 63.494 | 61.918 | 63.824 | 63.518 | 53.930 ± 8.68E-01 |
| MobileNetV2-0.75 | 70.260 | 68.866 | 68.870 | 68.884 | 68.873 ± 9.45E-03 | 67.840 | 68.866 | 69.380 | 67.696 | 69.608 | 69.304 | 58.958 ± 8.23E-01 |
| MobileNetV2-1 | 73.120 | 71.748 | 71.748 | 71.734 | 71.740 ± 7.21E-03 | 70.688 | 71.738 | 72.158 | 70.742 | 72.392 | 71.948 | 61.382 ± 7.27E-01 |
| MobileNetV2-1.4 | 75.844 | 74.856 | 74.824 | 74.832 | 74.837 ± 1.67E-02 | 73.692 | 74.856 | 75.122 | 73.798 | 75.396 | 74.970 | 63.978 ± 7.17E-01 |
| RegNetX-200M | 68.646 | 66.866 | 66.858 | 66.833 | 66.852 ± 1.72E-02 | 66.048 | 66.866 | 67.470 | 66.014 | 67.786 | 67.466 | 57.380 ± 7.66E-01 |
| RegNetX-400M | 72.220 | 70.791 | 70.772 | 70.776 | 57.780 ± 1.00E-02 | 69.982 | 70.792 | 71.204 | 69.756 | 71.520 | 71.348 | 60.659 ± 7.40E-01 |
| RegNetX-600M | 73.942 | 72.888 | 72.896 | 72.856 | 72.880 ± 2.12E-02 | 71.940 | 72.888 | 73.340 | 71.784 | 73.676 | 73.392 | 62.433 ± 7.98E-01 |
| RegNetX-800M | 75.246 | 74.338 | 74.332 | 74.330 | 74.333 ± 4.16E-03 | 73.458 | 74.338 | 74.786 | 73.394 | 75.052 | 74.718 | 63.679 ± 7.08E-01 |
| RegNetX-1.6G | 77.272 | 76.570 | 76.566 | 76.562 | 76.566 ± 4.00E-03 | 75.858 | 76.570 | 77.008 | 75.728 | 77.194 | 76.852 | 65.603 ± 6.11E-01 |

which—as expected—reduces half information. Another major difference comes from the third-party resize libraries. As shown in Fig. 2b and 2g, Pillow and OpenCV have different rounding mechanisms. As a result, they output totally different reversed images. In bicubic interpolation, Pillow and OpenCV also share opposed directions of gradients in color. In practice, the effect of resize operation on real-world images is not as obvious as our example here, however, it is enough to cause wrong predictions for a trained neural network.

## 4 THE IMAGENET-S ROBUSTNESS BENCHMARK

### 4.1 ImageNet-S Design

To benchmark the robustness of this unique systematic noise, we aim to build an ImageNet-S dataset. This dataset consists of 3 commonly used decoder types and 6 commonly used resize types. For decoder, we include the implementation from Pillow [29], OpenCV [3] and FFmpeg [27]. For resize operation, we include nearest, cubic and bilinear interpolation modes from both OpenCV and Pillow tools.

To verify the effect of decoding and resize separately, we must fix the decoding method when evaluating the resize and vice versa. In this paper, we set Pillow bilinear mode as the default resize method when testing the decoding robustness. This setting is also the default training setting in PyTorch [21] official code example. Similarly, we set Pillow as the default image decoding tool (same as PyTorch).

We provide not only the validation set of ImageNet with different decoding and resize methods, but also the training set to expose how different decoding and resize methods influence the process of training on a specific model. To enable reproducibility, we save each image file after decoding and resize it as a $3 \times width \times height$ matrix in a .npy file instead of JPEG. According to the commonly used transform on ImageNet [7] train set and test set, we provided two kinds of preprocessing for training and testing respectively. For training, we provide a matrix of an image after random resize crop to $3 \times 224 \times 224$. While for testing, we provide a matrix of an image after the process of resizing to $3 \times 256 \times 256$ then applied a center crop to $3 \times 224 \times 224$.

### 4.2 Dataset Generator

We provide the generator of ImageNet-S on GitHub[3], so that the community can generate system noise dataset and test the robustness of their own models. The basic usage of it is as follows:

```
from imagenet_s_gen import ImageTransfer

ImageTransfer(root_dir='./images/val',
              meta_file='./images/meta/val.txt',
              save_dir='./dataset-decoder-resize',
              decoder_type='pil',
              transform_type=val,
              resize_type='pil-bilinear')
```

## 5 EXPERIMENTS

Our experiments consist of four parts. First, we evaluate the robustness of existing pre-trained ImageNet models. These models all use the standard [Pillow decoder and Pillow bilinear resize] to train. Second, we study whether the commonly adopted $L_2$ or $L_\infty$ [5] robust models can defense our ImageNet-S case. Last, we study the training choice of decoding and resize to compare the results on ImageNet-S.

### 5.1 Network Architecture Experiments

In this section, we test pre-trained neural network models from different architecture to find out the relationship between architecture and systematic noise robustness. Additionally, we aim to find whether the complexity (e.g. FLOPs of the model) of a model can improve the robustness or not. We select architecture from ResNet [14] (including 18, 34, 50, 101 layers models), MobileNet-V2 [23] (including width multiplier 0.5, 0.75, 1 and 1.4× models) and RegNetX [22] (including FLOPs 200, 400, 600, 800, 1600M models). We summarize the results in Table 2 and Figure 3. As can be seen, we find several results: (1) Decoding corruptions are much less obvious than resize corruptions. In the decoding test, we find OpenCV, Pillow and FFmpeg have similar accuracy, with only < 0.01% accuracy difference. NVIDIA DALI, however, has a moderately different decoding process than these 3 libraries, which generally causes 1% accuracy difference. For resize method, we observe a relatively large gap. The biggest gap for resize operation comes from the nearest neighbor and bilinear interpolation, which generally costs 1 ∼ 2% accuracy gap. (2) Different architecture has different robustness, even with similar FLOPs. (3) For models in the same architecture, bigger FLOPs usually result in better robustness in different resize methods. (4) There is no clear relationship between FLOPs and the robustness of different decoders.

---
[3]https://github.com/TheGreatCold/Imagenet-S

Table 3: Top-1 accuracy of various robust ResNet-50 on ImageNet-S.

| Robust Type | $\epsilon$ | Decode | | | | | Resize | | | | | | |
|---|---|---|---|---|---|---|---|---|---|---|---|---|---|
| | | DALI | OpenCV | Pillow | FFmpeg | Mean ± Std. | Pillow-N | Pillow-L | Pillow-C | OpenCV-N | OpenCV-L | OpenCV-C | Mean ± Std. |
| None | 0 | 75.026 | 75.616 | 75.600 | 75.622 | 75.613 ± 1.14E-02 | 70.626 | 75.616 | 75.502 | 70.624 | 74.266 | 72.796 | 62.780 ± 2.27E+00 |
| $L_2$ | 0.01 | 75.058 | 75.386 | 75.392 | 75.412 | 75.397 ± 1.36E-02 | 71.490 | 75.386 | 75.240 | 71.578 | 74.442 | 73.312 | 63.067± 1.74E+00 |
| $L_2$ | 0.03 | 75.196 | 75.538 | 75.526 | 75.538 | 75.534 ± 6.93E-03 | 71.614 | 75.538 | 75.414 | 71.562 | 74.560 | 73.300 | 63.145 ± 1.80E+00 |
| $L_2$ | 0.05 | 75.036 | 75.240 | 75.224 | 75.240 | 75.235 ± 9.24E-03 | 71.740 | 75.240 | 75.264 | 71.732 | 74.462 | 73.238 | 63.100 ± 1.63E+00 |
| $L_2$ | 0.1 | 74.532 | 74.558 | 74.556 | 74.560 | 74.558 ± 2.00E-03 | 71.390 | 74.558 | 74.508 | 71.344 | 73.914 | 72.806 | 62.649 ± 1.47E+00 |
| $L_2$ | 0.25 | 74.112 | 73.730 | 73.714 | 73.722 | 73.722 ± 8.00E-03 | 71.384 | 73.730 | 73.810 | 71.328 | 73.356 | 72.426 | 62.293 ± 1.13E+00 |
| $L_2$ | 0.5 | 73.190 | 72.476 | 72.456 | 72.460 | 72.464 ± 1.06E-02 | 70.366 | 72.476 | 72.776 | 70.422 | 72.308 | 71.436 | 61.400 ± 1.06E+00 |
| $L_2$ | 1 | 70.548 | 69.600 | 69.606 | 69.606 | 69.604 ± 3.46E-03 | 68.130 | 69.600 | 69.976 | 68.112 | 69.618 | 68.956 | 59.200 ± 8.02E-01 |
| $L_2$ | 3 | 63.172 | 61.202 | 61.194 | 61.198 | 61.198 ± 4.00E-03 | 60.726 | 61.202 | 61.876 | 60.664 | 61.664 | 61.386 | 52.504 ± 4.90E-01 |
| $L_2$ | 5 | 56.400 | 54.292 | 54.270 | 54.280 | 54.281 ± 1.10E-02 | 53.900 | 54.292 | 54.764 | 53.880 | 54.584 | 54.370 | 46.542 ± 3.57E-01 |
| $L_\infty$ | 0.5/255 | 73.878 | 73.246 | 73.254 | 73.250 | 73.250 ± 4.00E-03 | 71.040 | 73.246 | 73.376 | 71.102 | 73.048 | 72.082 | 61.987 ± 1.07E+00 |
| $L_\infty$ | 1/255 | 72.298 | 71.266 | 71.252 | 71.256 | 71.258 ± 7.21E-03 | 69.846 | 71.266 | 71.594 | 69.794 | 71.374 | 70.690 | 60.654 ± 7.88E-01 |
| $L_\infty$ | 2/255 | 69.366 | 68.346 | 68.350 | 68.344 | 68.347 ± 3.06E-03 | 67.426 | 68.346 | 68.752 | 67.400 | 68.562 | 68.096 | 58.370 ± 5.73E-01 |
| $L_\infty$ | 4/255 | 63.982 | 62.518 | 62.508 | 62.514 | 62.513 ± 5.03E-03 | 61.458 | 62.518 | 63.066 | 61.542 | 62.730 | 62.140 | 53.352 ± 6.49E-01 |
| $L_\infty$ | 8/255 | 54.854 | 53.120 | 53.106 | 53.124 | 53.117 ± 9.45E-03 | 52.508 | 53.120 | 53.508 | 52.532 | 53.386 | 52.990 | 45.436 ± 4.20E-01 |

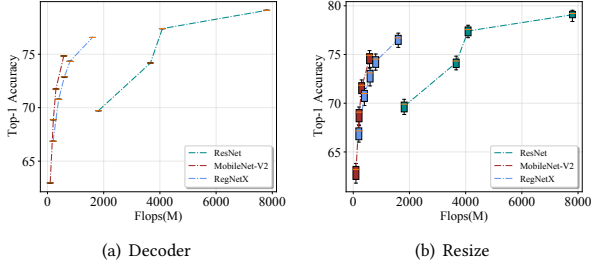

Figure 3: Architecture robustness of systematic noise.

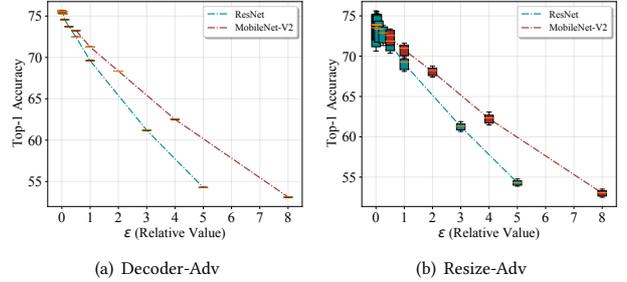

Figure 4: Architecture robustness of systematic noise.

## 5.2 Adversarial Experiments

Adversarial training is a defensive technique that gives model deceptive images during training, thus improving its generalization and robustness. Given a model $f_\Theta$ and an input image batch $X$ with the ground truth label $Y$, an adversarial example of $X_{adv}$ satisfies

$$f_\Theta(x_{adv}) \neq Y \quad s.t. \quad \|X - X_{adv}\|_p \leq \epsilon, \qquad (4)$$

where $\|\cdot\|_p$ is a distance metric. Commonly, $\|\cdot\|$ is measured by the $\ell_p$-norm ($p \in \{1, 2, \infty\}$).

Adversarial training is one of the most effective approaches to help deep learning models defend against adversarial examples. Thus, it is reasonable to use some adversarial models to test their robustness to our system noise. We download the pre-trained robust trained ResNet-50 [14] from this GitHub repository[4]. The ResNet-50 is trained with either $L_2$ or $L_\infty$ training method. The result is shown in Table 3. We find several interesting insights: (1) Neither of these two adversarial training methods improves the image decoder's robustness. (2) Both adversarial training methods can improve the robustness of different resize methods. The robustness increase as $\epsilon$ gets larger, but a large $\epsilon$ would harm the accuracy.

## 5.3 Robustness Enhancements

As aforementioned, we have verified that adversarial training can enhance resize method robustness, but it cannot help with the robustness on different decoders. This indicates that decoder noise does not belong to the $L_p$ norm perturbation. To overcome this problem, we need other techniques to enhance the systematic robustness.

*5.3.1 Mix Training.* To solve this problem, a natural way is to make the model "see" all kinds of decoders and resize methods during the training process. Based on this principle, we introduce *mix training* method to enhance the model's robustness on system noise.

The main process of mix training is to select decoder or resize method randomly instead of just using one kind of method during the whole process of training. The pseudocode of our algorithm is shown in algorithm 1.

---
[4] https://github.com/microsoft/robust-models-transfer

Table 4: Mix training on resize method.

| Train \ Test | Pillow-bilinear | Pillow-nearest | Pillow-cubic | OpenCV-nearest | OpenCV-bilinear | OpenCV-cubic | Mean | Std. |
|---|---|---|---|---|---|---|---|---|
| Pillow-bilinear | 76.572 | 72.168 | 76.512 | 72.090 | 75.346 | 74.072 | 74.460 | 2.02E+00 |
| Pillow-nearest | 74.872 | 75.988 | 75.548 | 75.970 | 76.002 | 76.056 | 75.739 | 4.63E-01 |
| Pillow-cubic | 76.312 | 72.828 | 76.596 | 72.876 | 75.810 | 74.666 | 74.848 | 1.68E+00 |
| OpenCV-nearest | 74.818 | 76.298 | 75.474 | 76.092 | 76.082 | 76.192 | 75.826 | 5.71E-01 |
| OpenCV-bilinear | 75.840 | 75.268 | 76.446 | 75.248 | 76.682 | 76.436 | 75.987 | 6.29E-01 |
| OpenCV-cubic | 76.194 | 72.812 | 76.510 | 72.940 | 75.736 | 74.818 | 74.835 | 1.62E+00 |
| mix | 76.154 | 75.876 | 76.344 | 75.786 | 76.444 | 76.330 | **76.156** | **2.70E-01** |

**Algorithm 1:** Mixed training for improving robustness on systematic noise.

**Input:** Resize set $\mathbb{RS}$; Decoder set $\mathbb{D}$; Model to train
Set Pillow-bilinear as default Resize;
Set Pillow as default Decoder;
**for** all $j = 1, 2, \ldots, T$-iteration in training **do**
  **if** use mix-decoder strategy **then**
    Randomly sample a Decoder from $\mathbb{D}$;
  **if** use mix-resize strategy **then**
    Randomly sample a Resize function from $\mathbb{RS}$;
  Calling API to load the images from ImageNet-S according to the Decoder type and Resize type;
  Model Optimization.
**return** An optimized robust model for systematical noise.

Table 5: Mix training on decoder.

| Train \ Test | Pillow | OpenCV | FFmpeg | Mean | Std. |
|---|---|---|---|---|---|
| Pillow | 76.430 | 76.426 | 75.310 | 76.055 | 6.45E-01 |
| OpenCV | 76.510 | 76.510 | 75.368 | 76.126 | 6.56E-01 |
| FFmpeg | 75.730 | 75.664 | 76.318 | 75.904 | 3.60E-01 |
| mix | **76.53** | **76.524** | **76.414** | **76.489** | **6.53E-02** |

To test the effect of mix training, we set up the following experiment. We use ResNet50 as the base model of this experiment. To comprehensively demonstrate the training effect, we train single decoding and resize as well as our mix training models. We set the default decoder as Pillow and the default resize method as Pillow bilinear when conducting ablation studies on resize method or decoder, respectively. Then we use top-1 accuracy as well as their mean and standard deviation as assessments.

The results of this experiment are shown in Table 5 and Table 4. From these tables, we can know that: (1) The model has a better performance (usually the best) when we train and test using the same decoder and resize method. (2) There is a big gap when we train and test the same model using different methods. For example, the accuracy of ResNet50 trained by Pillow bilinear method is about 2.4% lower than that trained by OpenCV bilinear on the test set which is resized by OpenCV cubic. While, on the whole, the gap between different decoders is smaller than that between different resize methods. (3) Mix training can improve the robustness of a model on system noise greatly without hurting the clean accuracy. The *Std.* using mix training drop from 0.36 to 0.0653 on decoder experiment, and drop from 0.463 to 0.270 on resize experiment. Meanwhile, it can maintain the model's accuracy at about 76%. As a contrast, the same ResNet50 model using $L_\infty - Robust$ adversarial training drop the *Std.* from 1.07 to 0.420 by paying 19.2% drop of clean accuracy.

## 6 CONCLUSION

In this paper, we introduced what is to our knowledge the first dataset, ImageNet-S, for real-world robustness from systematic noise. We focused on a problem that exists in the process of practical application, but it is easy to be ignored in practice. We finished lots of experiments that do show that the difference between different decoders and resize methods can influence on model's accuracy. Therefore, we used mean and standard deviation which are used to measure the robustness of systematic noise. With this matrix, we did experiments and found the relation between architecture, adversarial, and the robustness of a model. What's more, we found a way called mix training to enhance the system noise robustness without hurting the clean accuracy.